# Target Height Estimation Using a Single Acoustic Camera for Compensation in 2D Seabed Mosaicking

Xiaoteng Zhou, Yusheng Wang, and Katsunori Mizuno

***Abstract*— This letter proposes a novel approach for compensating target height data in 2D seabed mosaicking for low-visibility underwater perception. Acoustic cameras are effective sensors for sensing the marine environments due to their high-resolution imaging capabilities and robustness to darkness and turbidity. However, the loss of elevation angle during the imaging process results in a lack of target height information in the original acoustic camera images, leading to a simplistic 2D representation of the seabed mosaicking. In perceiving cluttered and unexplored marine environments, target height data is crucial for avoiding collisions with marine robots. This study proposes a novel approach for estimating seabed target height using a single acoustic camera and integrates height data into 2D seabed mosaicking to compensate for the missing 3D dimension of seabed targets. Unlike classic methods that model the loss of elevation angle to achieve seabed 3D reconstruction, this study focuses on utilizing available acoustic cast shadow clues and simple sensor motion to quickly estimate target height. The feasibility of our proposal is verified through a water tank experiment and a simulation experiment.**

## I. Introduction

The ocean is regarded as a critical region supporting the sustainable development of humanity in the future. Currently, industrial projects such as offshore wind farm construction, clean fuel exploration, and deep-sea mining are actively being explored. However, these applications are often located in low-visibility marine environments, as shown in Fig. 1, where effective perception of the seabed environments has been a main challenge [1]. In low-visibility marine environments, such as in dark or turbid conditions, dominant optical sensing systems often fail due to insufficient lighting and interference from suspended particles. In contrast, imaging sonar, as another primary underwater visual sensor, exhibits strong robustness in low-visibility environments [2]. Sonar imaging does not rely on light sources and can operate in complete darkness, thus enabling day and night surveys. Additionally, the longer wavelength of sound waves allows them to bypass particles, which enables sonar to work in turbid water. It is widely acknowledged that improving underwater perception in low-visibility conditions is crucial for the development of advanced marine robots, as it significantly enhances their navigation and survey capabilities in harsh environments.

*This research was performed by the Environment Research and Technology Development Fund (JPMEERF20231004) of the Environmental Restoration and Conservation Agency of Japan. This work was also supported by JST SPRING, Grant Number JPMJSP2108.

Xiaoteng Zhou and Katsunori Mizuno are with the Graduate School of Frontier Sciences, The University of Tokyo, Kashiwa 277-8583, Japan.

Yusheng Wang is with the Graduate School of Engineering, The University of Tokyo, Tokyo 113-8654, Japan.

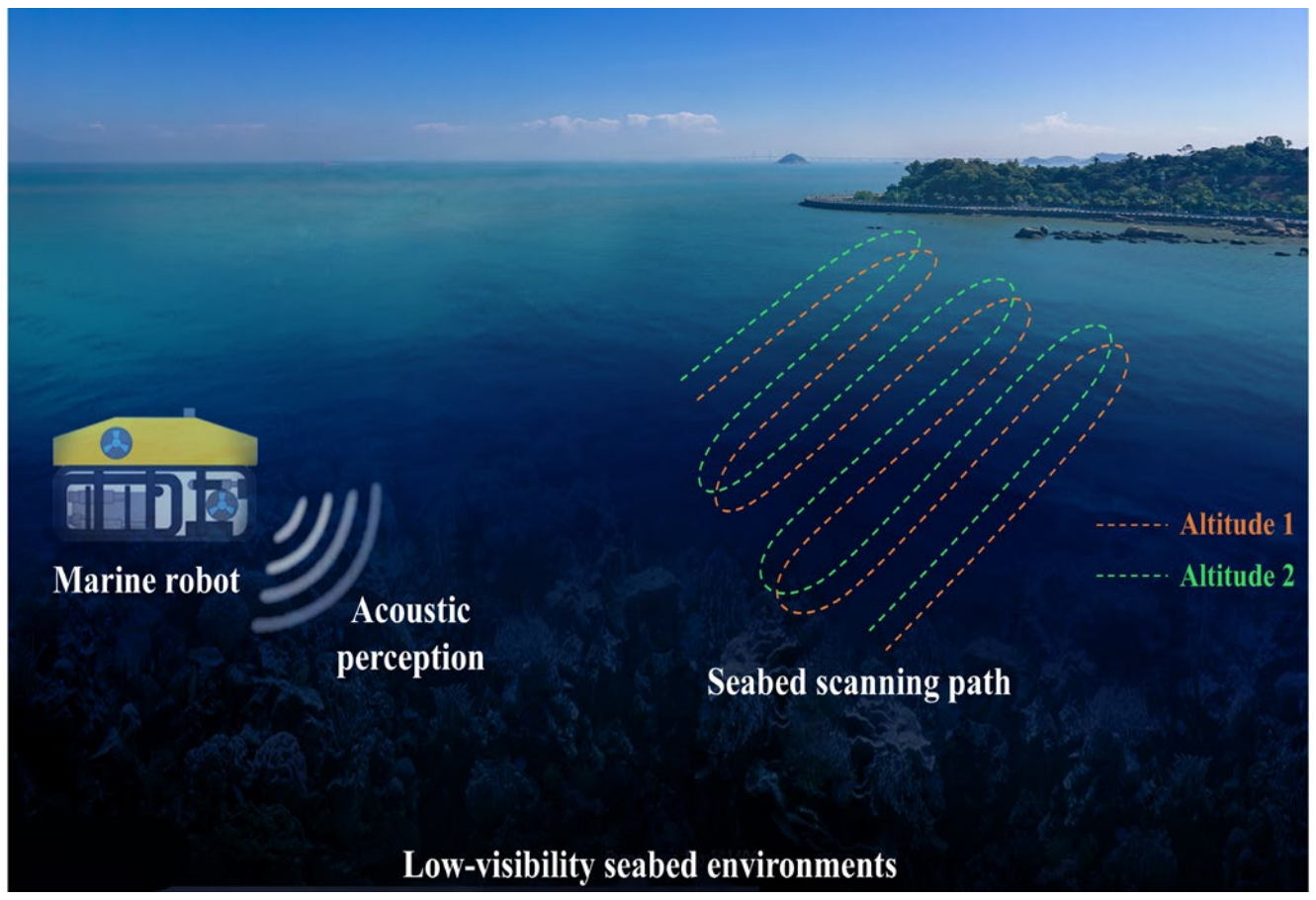

Figure 1. Seabed investigation in low-visibility marine environments.

Acoustic cameras, also known as active forward-looking sonar (FLS), have garnered extensive attention due to their robust imaging capability in low-visibility conditions [3]. Compared to commonly used side-scan sonar and mechanical scanning sonar, the imaging mode of acoustic cameras more closely resembles that of optical cameras [4], allowing researchers to intuitively understand seabed environments. Moreover, the acoustic camera is compact and lightweight, making it flexible to install on marine robots for underwater surveys [5]. Furthermore, due to the high-resolution imaging, acoustic cameras are suitable for seabed mosaicking. In marine engineering applications, mosaicking techniques are used to stitch local ocean scene images into comprehensive panoramas, which help to visually display seabed topography and the distribution of objects [5]. Fig. 2 presents a seabed mosaicking generated using an acoustic camera. Additionally, acoustic cameras can acquire real-time environmental visual data and provide distance measurements that optical cameras cannot achieve, thereby offering crucial support for marine robots in navigation and obstacle avoidance within complex underwater environments [6].

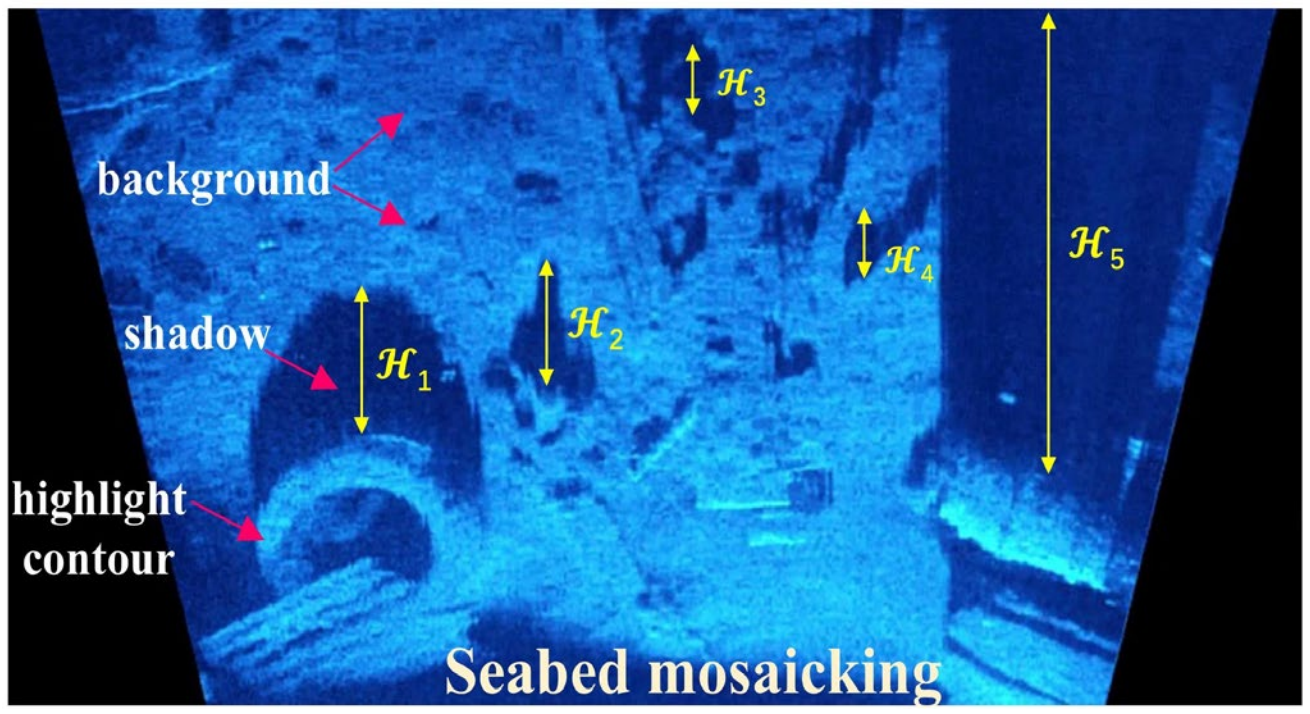

Figure 2. Seabed mosaicking by an acoustic camera, with the data from [4].

However, most existing seabed mosaicking pipelines [5], [7], [8] can only produce two-dimensional (2D) mapping results, primarily due to the loss of target height information during acoustic imaging. To tackle this issue, several studies have utilized three-dimensional (3D) reconstruction pipelines to restore the 3D structure of targets and extract critical height data [9]-[12]. Although these methods have achieved initial success, their complexity has limited their application in underwater real-time surveys to a certain extent. Secondly, these methods often require sufficient detection viewpoints to capture overlapping target images, which is challenging to achieve in real underwater surveys. Although these methods have high accuracy in reconstructing target shapes, precise replication of objects is not a pressing need for marine robots unless tasks such as grasping are involved. In practice, the 2D seabed mosaicking combined target height estimation generally meets the requirements for survey and navigation.

To support practical engineering applications, this article proposes a novel research approach that utilizes shadow clues in acoustic images to directly estimate the height of targets and compensates this information into 2D seafloor panoramas, as shown in Fig. 2, thereby enabling better seabed investigations.

## II. Related Works

In acoustic camera image analysis, key visual cues include highlight contour, shadow, and background, as shown in Fig. 2. Previous research has utilized the highlight contour to estimate target height, establishing height estimation models based on the movement characteristics of the highlight area [11], [13]. These methods rely on the field of view (FOV) of the acoustic camera to accurately capture targets. However, this process faces several challenges. Firstly, there may be errors in the sonar beam emission angle, resulting in an inability to accurately define the FOV. Secondly, sonar systems are often affected by sidelobe effects, where echoes from non-main beam directions interfere with the accuracy of target capture. Finally, sound waves experience scattering and attenuation during propagation, which can collectively reduce the target capture accuracy. Wang et al. [10] proposed a novel acoustic camera-based dense mapping method that combines a rotator and a voxel model to achieve the construction of seabed maps. By constructing an inverse sensor model of the acoustic camera, the method can update the probability of each voxel and generate local 3D maps from multiple views. However, the resolution of the mapping results output by this method is not enough, and due to its computational complexity, it is not currently suitable for rapid seabed surveys. The discussion of other types of 3D reconstruction methods is not included in this paper, as the research motivation focuses on achieving fast and robust target height estimation.

To meet the requirements of robustness and rapidity in seabed survey, this study proposes a target height estimation method based on target cast shadows. This is a pragmatic decision because, whether objects are structured or unstructured, shadow regions will appear behind them and usually occupy a large portion of the image content. Additionally, shadows, as an effective visual cue in acoustic camera images, can significantly reduce the reliance on complex handcrafted feature operators, thereby lowering computational complexity. In short, the method proposed in this study only needs to scan the seabed at different altitudes to achieve high-resolution seabed mosaicking and accurate target height measurement, as shown in Fig. 1.

Using cast shadows to estimate target height is a classic approach in acoustic image processing [14]. However, this method is primarily suitable for independent survey scenarios and has not yet been effectively integrated with dynamic seafloor mosaicking. Therefore, the practical application value of estimating the target height alone is relatively limited. In other words, if the target height can be accurately estimated and effectively integrated into the 2D seafloor mosaicking without significantly increasing survey costs, this method will hold considerable practical value. The method designed in this study emphasizes practicality, achieving a reasonable balance between simple 2D representation and the complex 3D reconstruction of seabed targets. It can be regarded as a 2.5D recovery approach.

In summary, this letter presents a novel low-visibility seabed environment survey method based on a single acoustic camera. The method uses acoustic camera images to perform 2D seafloor mosaicking and makes full use of cast shadows in these sonar images to estimate seabed target height, thereby achieving 3D measurements of the seafloor environments. Compared to previous research, this study does not rely on complex sonar elevation modeling but instead conducts 3D seafloor surveys based on visual clues available in acoustic images. Furthermore, this pipeline is flexible and sub-steps can be optimized using machine vision algorithms. The contributions of this work are listed as follows:

1) Transform the target height estimation issue into the analysis of the physical phenomenon of cast shadows, thereby reducing reliance on complex sonar elevation modeling.

2) By using a single camera to scan the seabed at different altitudes, high-resolution seabed mosaicking and target height information can be obtained.

3) Develop a balanced approach to reliably acquire critical 3D target data required for seabed surveys with minimal additional detection costs.

4) This study provides a new reference for rapid seabed information acquisition of marine robots and offers technical support for future underwater autonomous operations.

The remainder of this article is organized as follows. Section III provides a detailed description of the method. Section IV introduces the experimental setting. Section V presents the results and objective evaluation. Some discussion is provided in Section VI. Finally, the conclusions and future works are presented in Section VII.

## III. Methodology

In surveys of seabed environments, most targets are unstructured. Research on information acquisition from unstructured targets can elucidate fundamental principles that are equally applicable to the analysis and processing of structured targets. The research pipeline is illustrated in Fig. 3.

This study primarily focuses on the height estimation of seabed targets, as the length and width of the targets can be directly obtained through 2D seabed mosaicking. For a comprehensive discussion on the seabed mosaicking process based on acoustic camera images, please refer to [15].

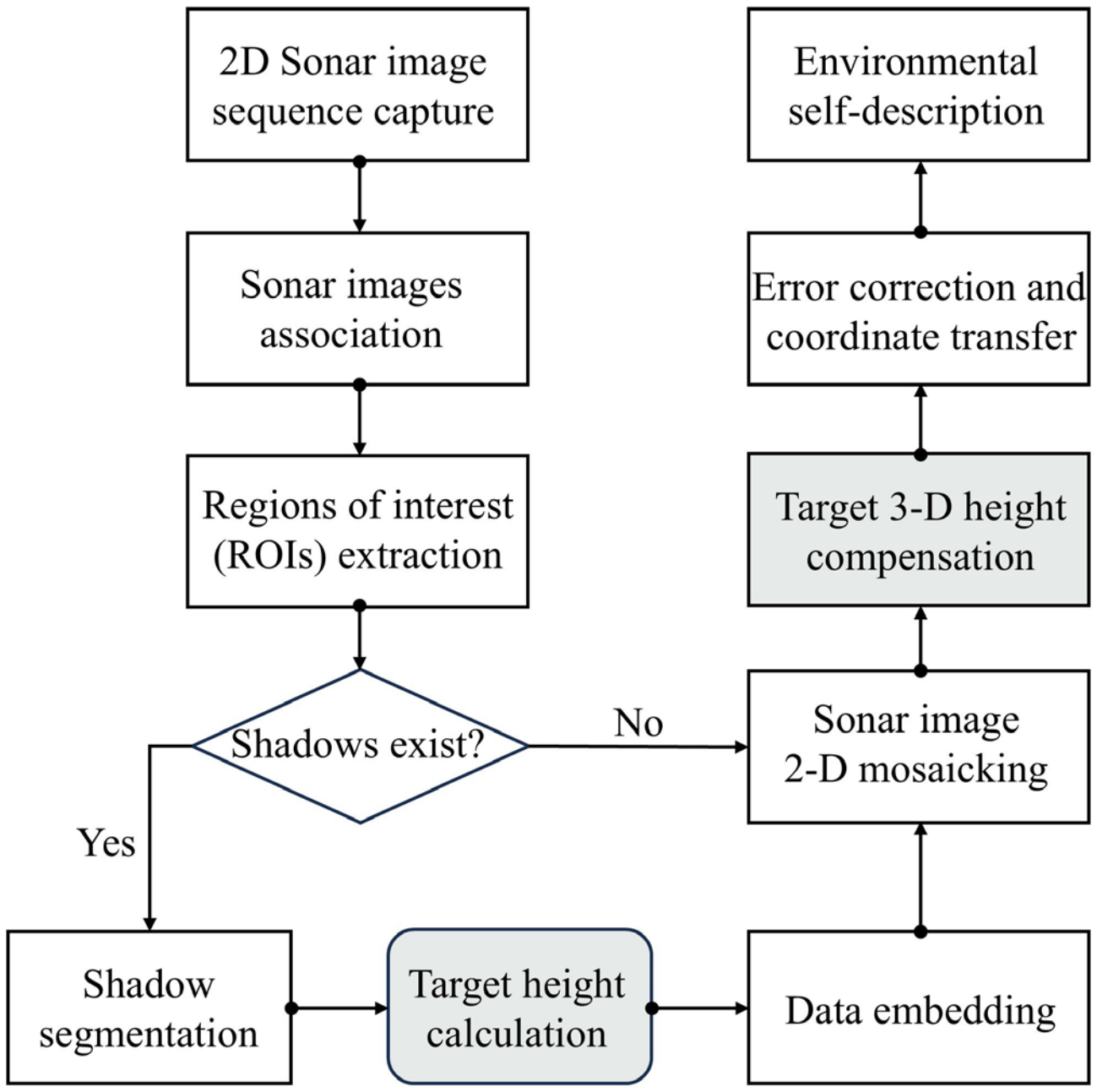

Figure 3. The pipeline of the overall research.

## A. Acoustic Camera Imaging Mechanism

The acoustic camera is a multi-beam FLS, which generates acoustic digital imaging by emitting multiple sound beams forward and measuring the intensity and travel time of the returning beams. Due to the distribution of the sound beam along the azimuth direction, the FOV of the acoustic camera is fan-shaped or wedge-shaped. The 3D point $(r,\theta,\varphi)$ in polar coordinate system could be projected to a 2D point $(r,\theta)$ in the zero-elevation imaging plane, as displayed in Fig 4. Therefore, it can be observed that the sonar image produced by the acoustic camera is essentially a 2D projection of the 3D physical world. Due to the ambiguity in the elevation angle during the imaging process, the Z dimension, which contains height data, is absent from the acoustic camera images [16].

$$\left[E_X E_Y E_Z\right]^{\mathrm{T}} = \left[r\cos\varphi\cos\theta \quad r\cos\varphi\sin\theta \quad r\sin\theta\right]^{\mathrm{T}} \qquad (1)$$

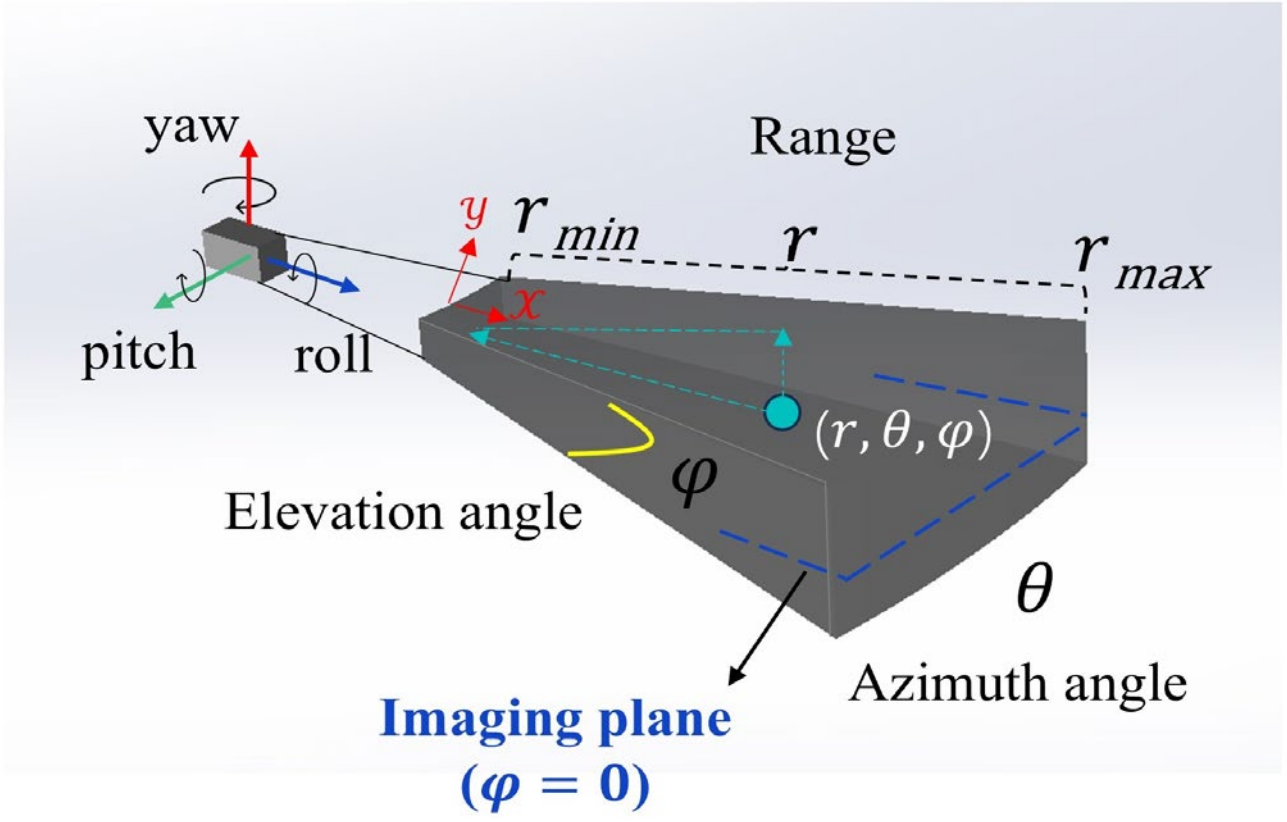

Figure 4. Acoustic camera imaging model.

Fig. 5 presents a raw acoustic camera image, labeled with explanatory annotations. This image was captured during this investigation and is a frame from an acoustic image sequence. The raw seabed survey data acquired by sonar is presented in the form of a time series and is often affected by noise. Therefore, image preprocessing steps such as association and denoising are typically required before downstream analysis.

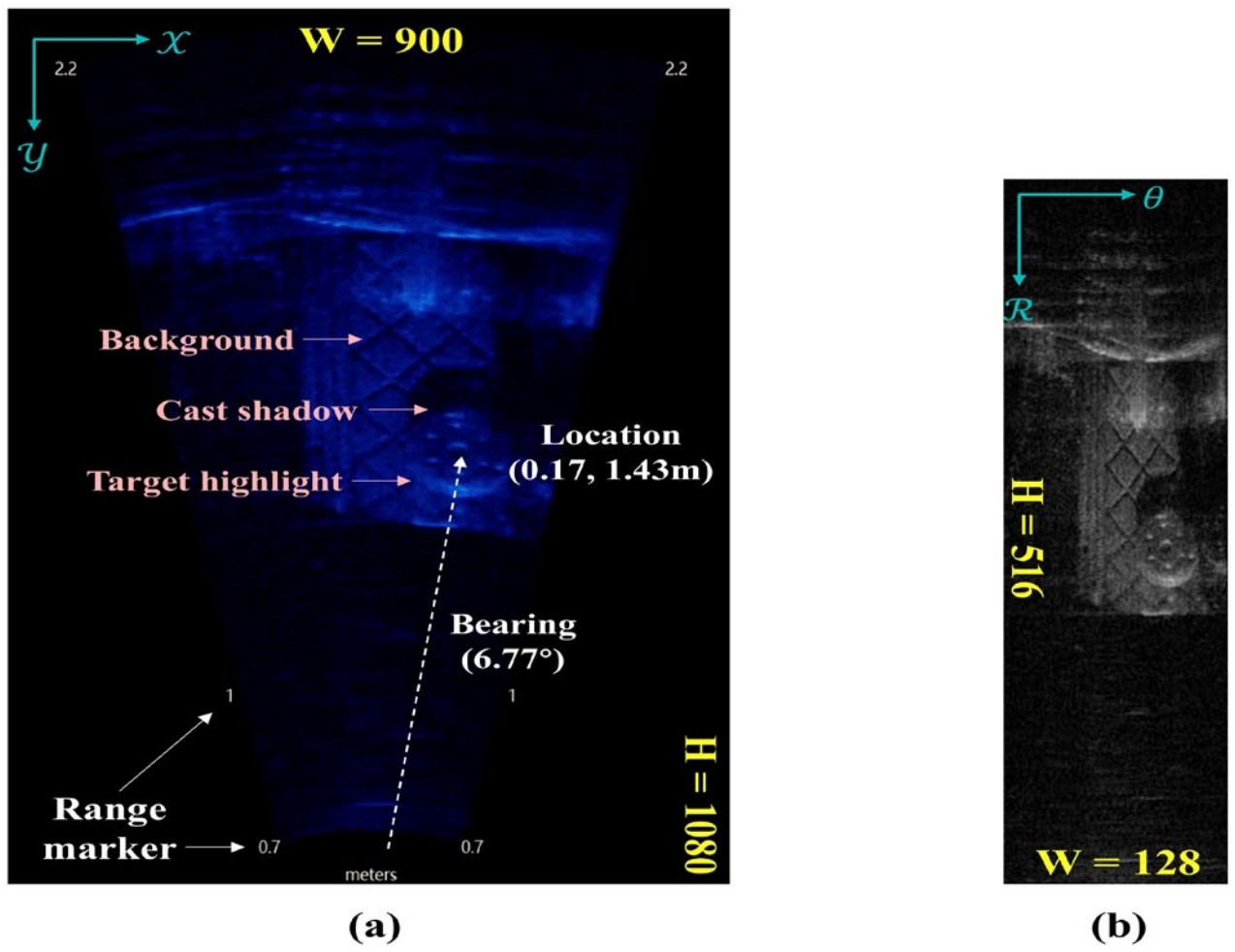

Figure 5. Rubber ring images captured by the acoustic camera. (a) In the Cartesian coordinate system, (b) in the polar coordinate system.

## B. Acoustic Camera Imagery Visual Clues Analysis

It can be seen from Fig. 5, that sonar images are typically composed of three major regions, namely, the target highlight area, the cast shadow area, and the background area on which target shadows are cast [14]. The target highlight area can be used to resolve the 2D information of the target, such as size, bearing, and location. The width of the cast shadow can be used to estimate the width of the target, while the length of the cast shadow mainly reflects the height of the target, rather than the length of the target. Cast shadows usually occupy a large portion of sonar images, and their spatial distribution exhibits distinct patterns, making them important visual clues.

## C. Impacts of Sonar Posture Adjustments on Cast Shadows

This study uses cast shadow to estimate the target height, and its distribution can be adjusted by the grazing angle of the sound beam. The grazing angle refers to the angle between the sonar wave beam and the target surface [12], as depicted in Fig. 6. Adjusting the posture of the acoustic camera (mainly altitude and pitch) will change its grazing angle, resulting in a deviation in the distribution of the cast shadow. This physical phenomenon can be used to infer the target height. In this work, we validated the proposed approach based on a simplified environmental assumption, primarily focusing on the flat seabed surface.

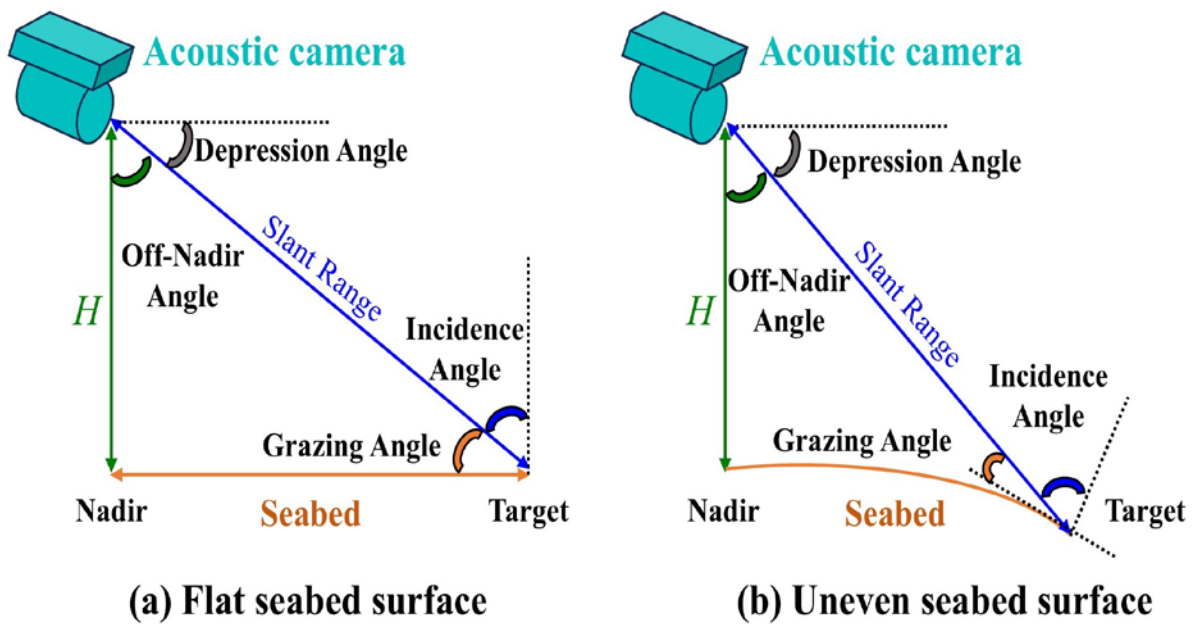

Figure 6. Grazing angle of the acoustic camera when scanning the seafloor.

As depicted in Fig. 7, adjusting the altitude of the acoustic camera will affect the length of the target cast shadow, while changes in the pitch angle will affect the location of the shadow within the sonar FOV. Both adjustments are essential: the former is used to determine the variation in shadow length for estimating the target height, while the latter ensures that the shadow is entirely within the FOV to reduce estimation errors. The pitch angle adjustment is not fixed but is fine-tuned based on the quality of the cast shadow distribution.

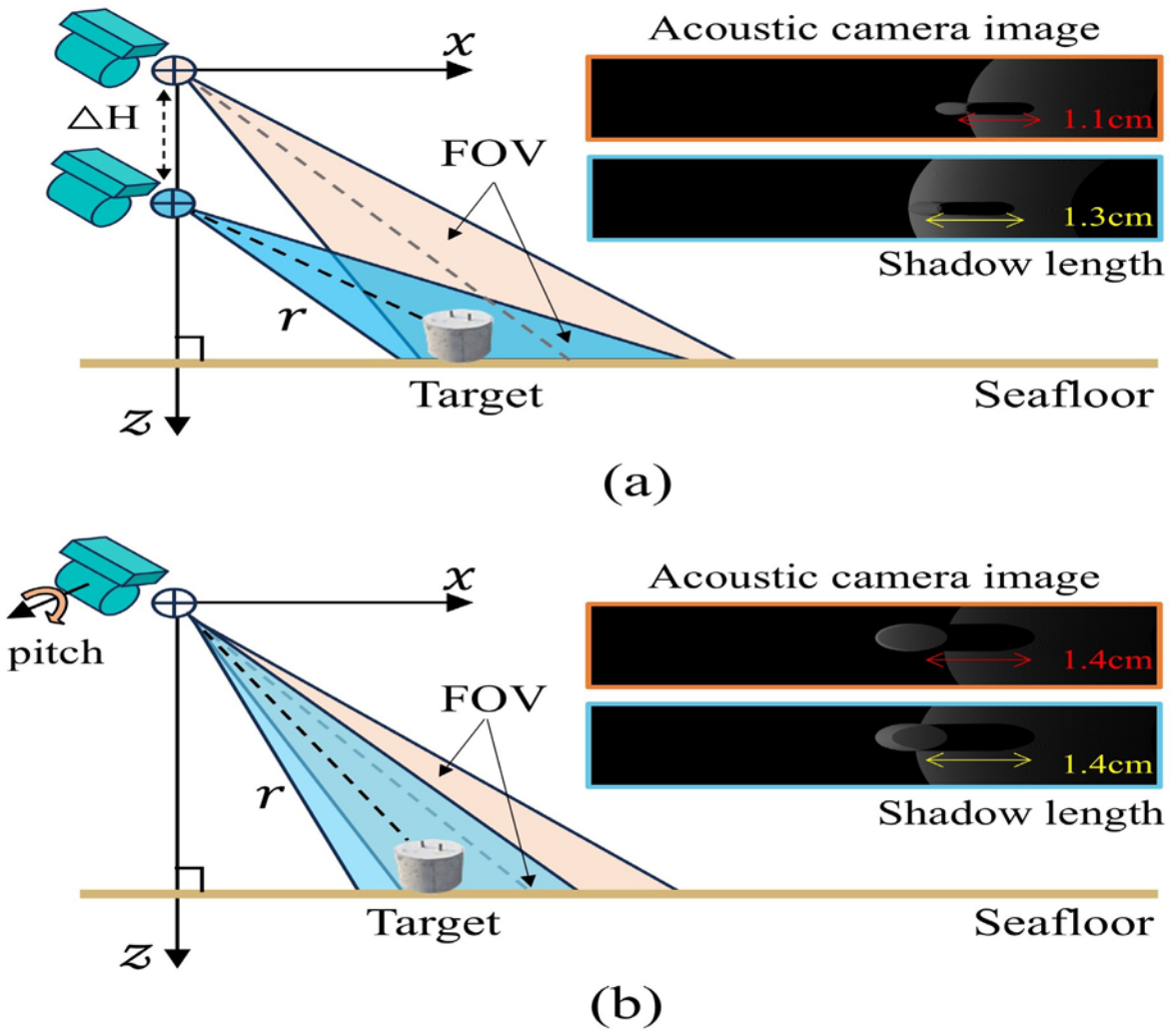

Figure 7. The impact of posture adjustments on the distribution of shadows.

Specifically, an acoustic camera is first used to vertically scan the seabed to obtain initial altitude data, as shown in Fig. 8(a). Then, by adjusting the altitude of the acoustic camera to affect the length of the cast shadow, the height of the target is calculated, as depicted in Figs. 8(b) and (c). During this process, if the cast shadow did not completely fall within the FOV (the gray area of each figure), the location of the cast shadow can be adjusted by modifying the pitch angle of the acoustic camera, as shown in Fig. 8(d) to (f). In this study, the posture adjustments of the acoustic camera involve only slight altitude and pitch adjustments, which are easy to achieve in underwater surveys. Furthermore, previous research has demonstrated that these two types of posture adjustments could significantly enhance the acoustic imaging quality [14].

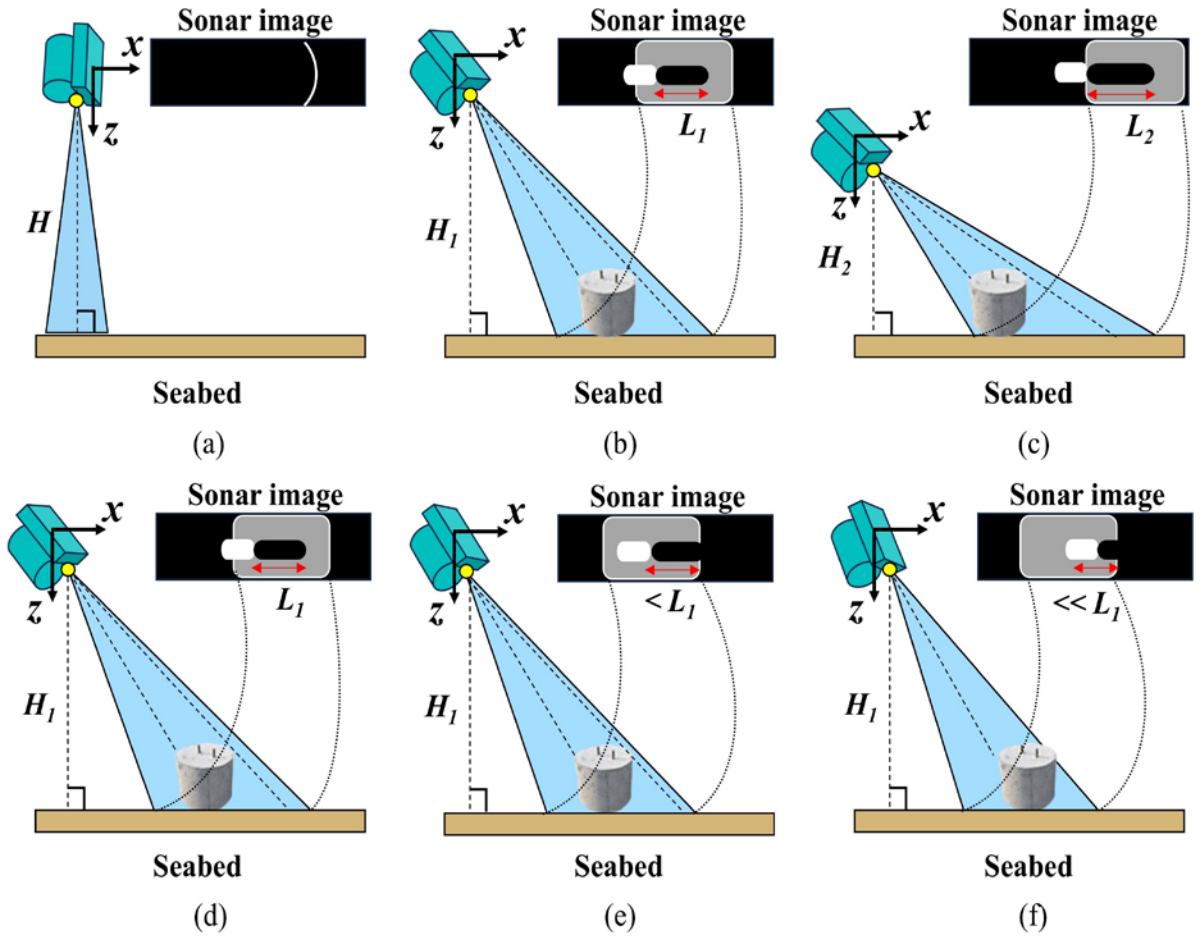

Figure 8. Two types of sensor posture adjustments are used to estimate target height. The design of this figure was inspired by work [10] and [12].

## D. Target Height Solving Model

When sound waves illuminate seabed targets at a specific grazing angle, the resulting image highlights the contours of these targets. The cast shadow area $B*E*$ is behind the highlight area $A*B*$, while the outer areas of the two correspond to the seabed background area, as shown in Fig. 9.

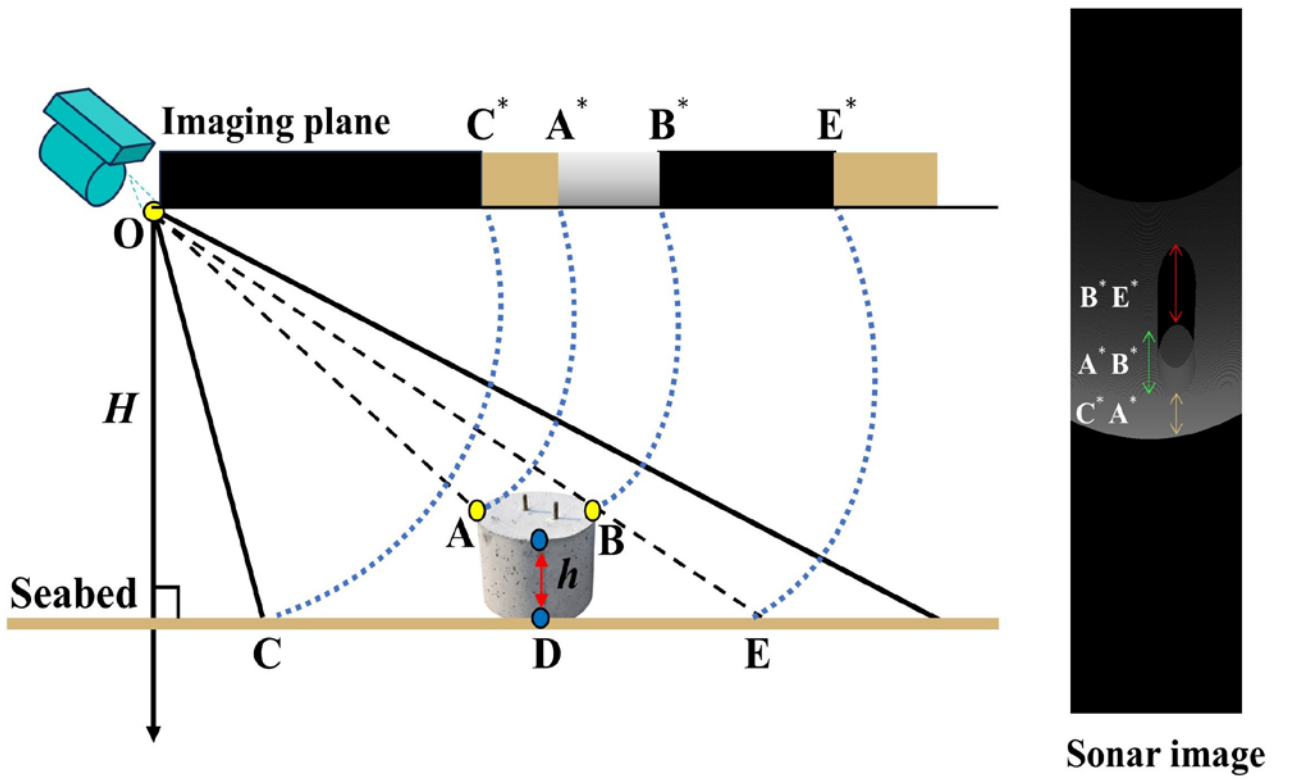

Figure 9. Acoustic imaging analysis of the target located on the seabed.

In actual surveys, the marine robot only needs to scan the seabed at different altitudes to obtain 2D mosaicking and the difference in shadow length of targets, as shown in Fig. 1. Then, combined with altitude variations and the triangulation principle [14], the height of seabed targets can be calculated, as illustrated in Fig. 10. During this process, it is not necessary to know the specific sonar altitude; it is sufficient to obtain the altitude variation that causes changes in cast shadow length. It is worth noting that using different altitudes to scan the seabed will not significantly increase the cost of survey, but will help further improve the accuracy of 2D seabed mosaicking [5].

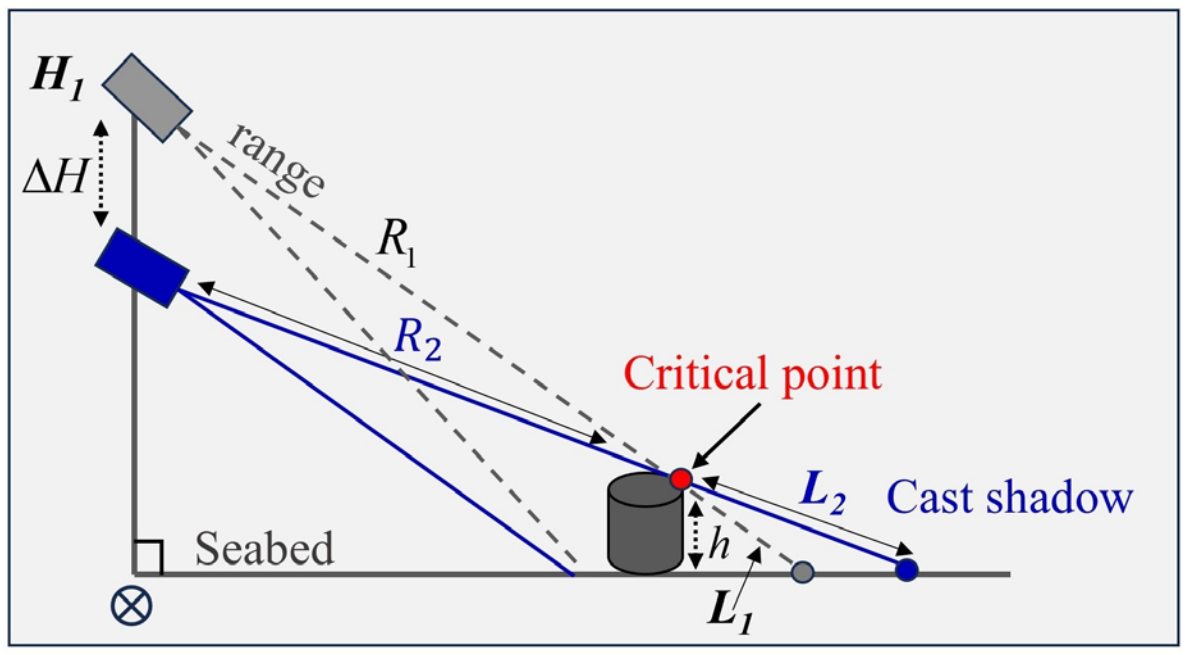

Figure 10. Diagram illustrating the method for estimating object height based on the altitude variation of the acoustic camera.

The equation for inferring the target height via cast shadow visual clues and acoustic camera movement is as follows:

$$h = \frac{\Delta H \cdot L_1 \cdot L_2}{(L_1 + R_1) \cdot L_2 - (L_2 + R_2) \cdot L_1} \quad (2)$$

where $h$ denotes the height of the seabed target, and $\Delta H$ denotes the sonar altitude variation. $L$ represents the pixel distance of the cast shadow. $R$ is the pixel distance from the shadow critical point to the detection baseline. Among these parameters, $\Delta H$ can be obtained from the depth data inside the acoustic camera, or it can be obtained through two initializations, as displayed in Fig. 8(a). The remaining parameters could be measured from the acoustic image.

The calculation of target height is detailed in Algorithm 1.

**Algorithm 1:** Height solving of targets on acoustic images

**Input:** Acoustic image pairs of targets at an altitude variation $(I_o, I_{o+\Delta H})$, custom thresholds $[m, n]$

**Output:** Target height set $H = \{h_1, h_2, h_3, \ldots, h_i, \ldots, h_k\}$

1 **For** acoustic image $I_o$ **do**

2 Locate target highlight areas $T = \{T_1, T_2, T_3, \ldots, T_i\}$

3 Locate acoustic shadow areas $S = \{S_1, S_2, S_3, \ldots, S_j\}$

4 **If** $T_i \subseteq \text{FOV} \wedge S_j \subseteq \text{FOV}$ **then**

5 derive pair set $P = \left\{ (T_i, S_j) \middle| m \le \frac{|T_i \cap S_j|}{|T_i|} \le n \right\}$

6 fit region critical line $L_R = \text{fit}(T_i, S_j)$

7 calculate the critical line pixel range $R_i$

8 calculate the shadow pixel range $L_j = length(S_j)$

9 **end**

10 Return $(R_i, L_j)$

11 For image $I_{o+\Delta H}$, repeat steps 2-8 to obtain $(R_i', L_j')$

12 Use Equation (2) to calculate the target height $h_i$

13 **Return** $H \leftarrow H \cup \{h_i\}$

## IV. Experimental Setting

### A. Water Tank and Sensor Specifications

This study validated the height estimation model through a real water tank experiment and a simulation experiment. Both experiments used the same target sizes and acoustic camera specifications, as detailed in Tables I and II.

TABLE I. Water Tank Parameters and Target Sizes

| Water tank parameters | |
|---|---|
| Length [cm] | 180 |
| Width [cm] | 60 |
| Height [cm] | 90 |
| Water depth [cm] | 85 |
| Material | Glass |

| Target sizes [L×W×H] | |
|---|---|
| 1 | 19 ×19× 2.8 cm |
| 2 | 10 × 5 × 3.9 cm |
| 3 | 5 × 4 × 4.7 cm |
| 4 | 9 × 4 × 3.0 cm |
| 5 | 6 × 5 × 2.8 cm |

TABLE II. Acoustic Camera Specifications

| Specification item | Value |
|---|---|
| Sonar hardware model | ARIS EXPLORER 3000 |
| Identification frequency [MHz] | 3.00 |
| Number of transducer beams | 128 |
| Width of each beam [Degree] | 0.25 |
| Azimuth angle [Degree] | 32.0 |
| Elevation angle [Degree] | 14.0 |

### B. Water Tank Experiment

This experiment was conducted in an indoor glass water tank, with the tank parameters detailed in Table I. Specifically, we utilize an acoustic camera to capture the target located below and continue detection while varying the altitude of the acoustic camera. The experimental layout is shown in Fig. 12, and targets captured by the sonar are shown in Figs. 5 and 11. Moreover, Fig. 14 presents the analysis of visual clues.

During the testing process, we precisely measured the 3D dimensions of all targets using a ruler and recorded the results in Table III. This study primarily focuses on estimating the height of seabed targets; therefore, we visualized the actual height of each target and used it as the ground truth (GT), as displayed in Fig. 15. To obtain the altitude difference, we use an internal depth gauge and a ruler to record synchronously.

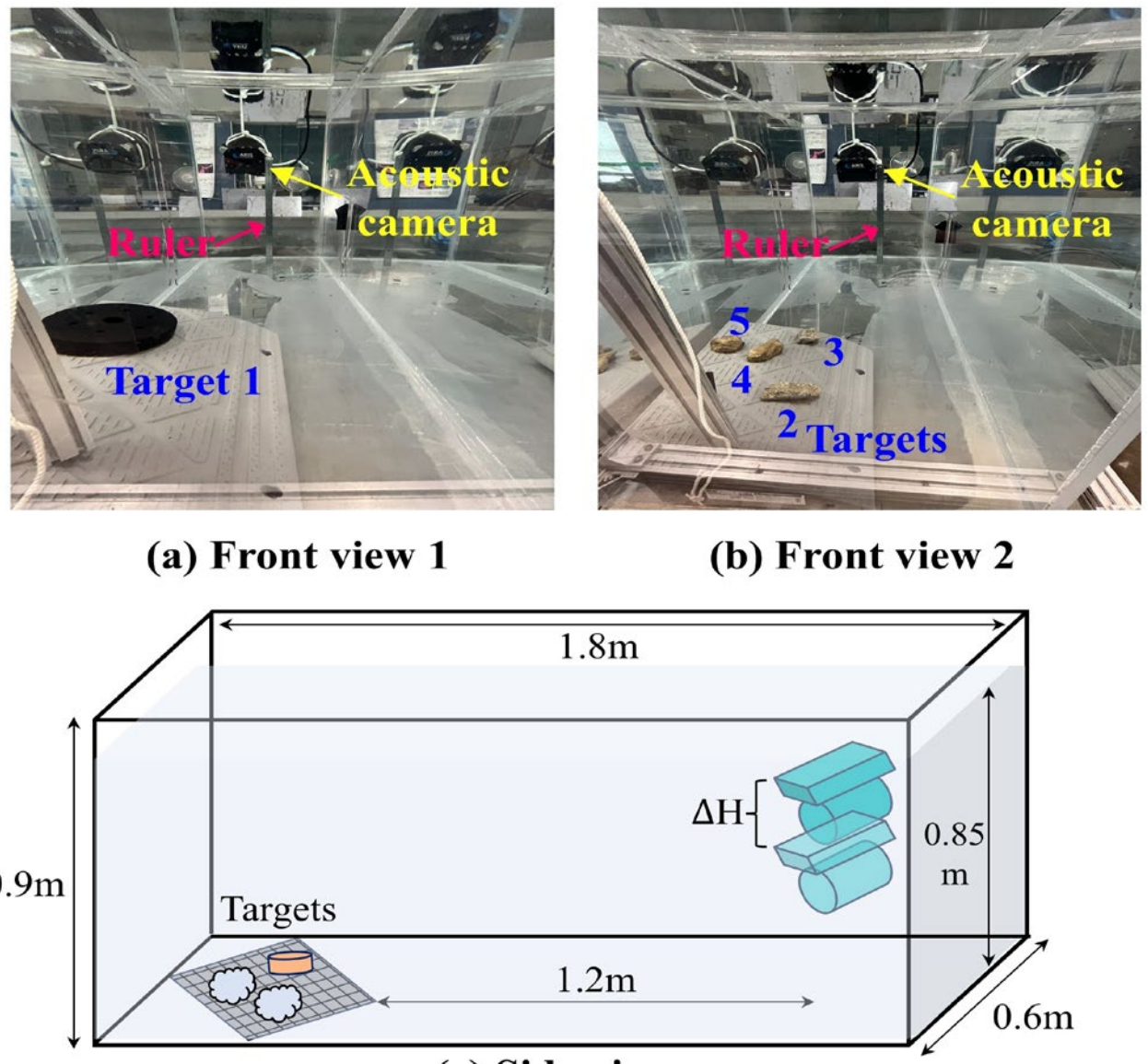

Figure 11. Water tank experimental layout in this study.

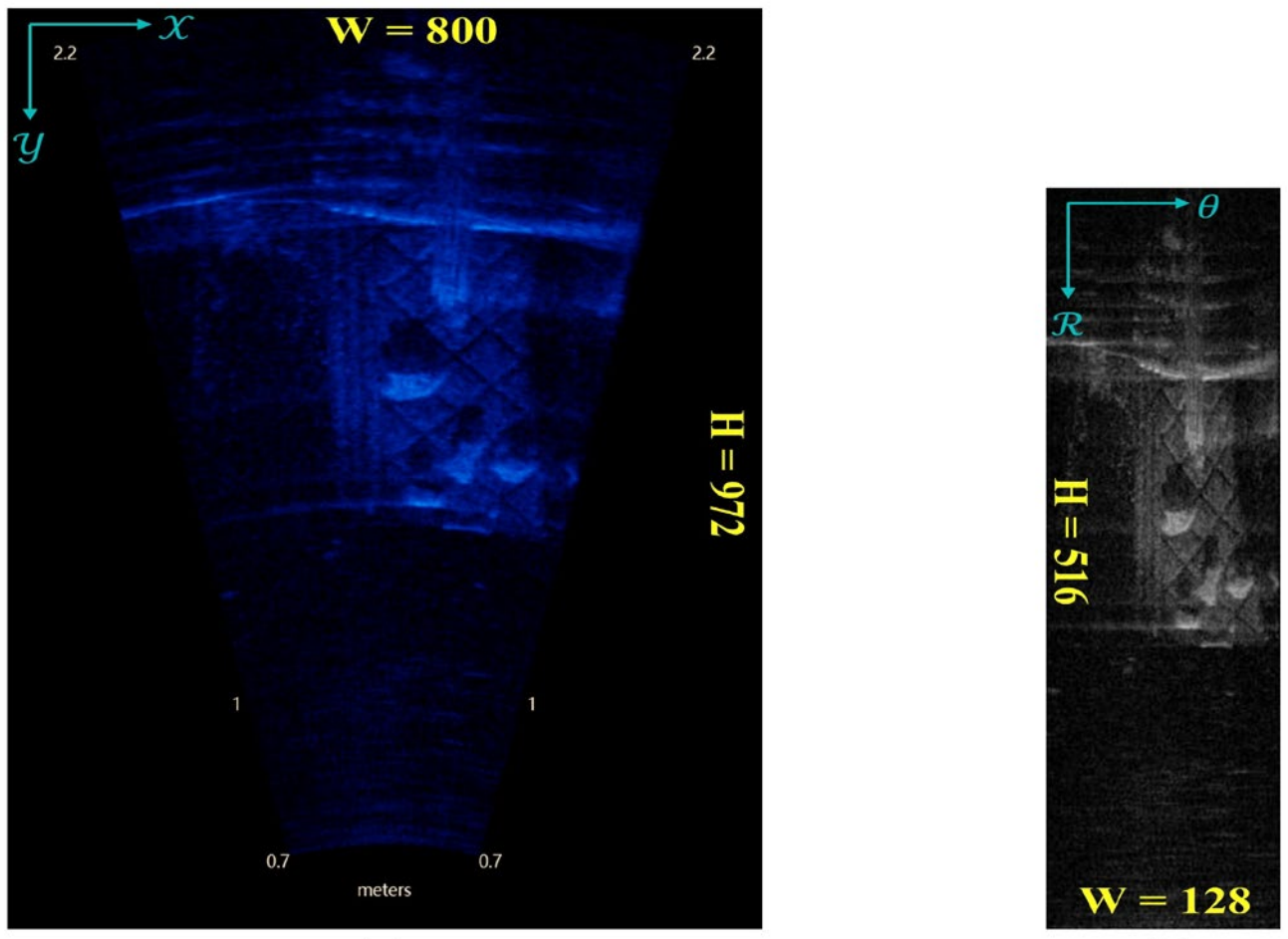

Figure 12. Stone images captured by the acoustic camera. (a) In the Cartesian coordinate system, (b) in the polar coordinate system.

### C. Simulation Experiment

The simulation environment is constructed using the open-source software Blender. A flashlight and a camera with the same pose are used to simulate an acoustic camera, serving as the emitter and receiver, respectively. For further details on the acoustic camera simulator, please refer to [16].

In the simulated scenario, the size of each target is consistent with those of the actual targets in the water tank experiment, and the sensor movement track is also maintained consistently. An overview of the simulated seabed scenario and the generated target acoustic images are displayed in Figs. 13 and 16.

The purpose of conducting a simulation experiment is to accurately calibrate and optimize the model, because in actual tests, the material properties of the target as well as external noise and artifacts may affect the features on the acoustic image, thereby reducing the accuracy of the target height estimation. In addition, simulation experiments can quickly evaluate the impact of various input parameters on the estimated results.

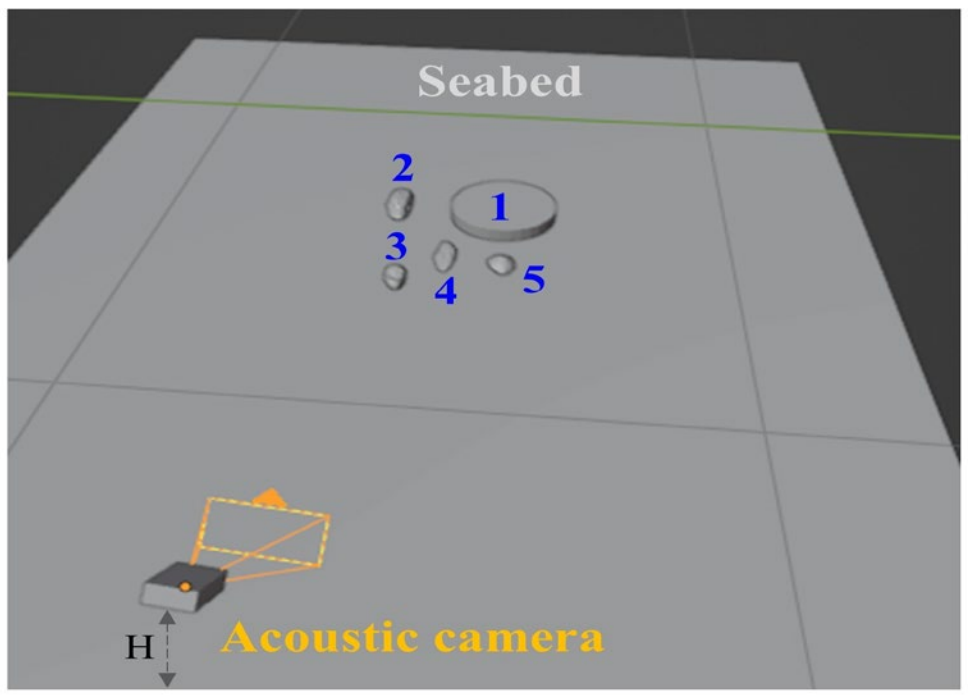

Figure 13. Survey scene generated by the acoustic camera simulator.

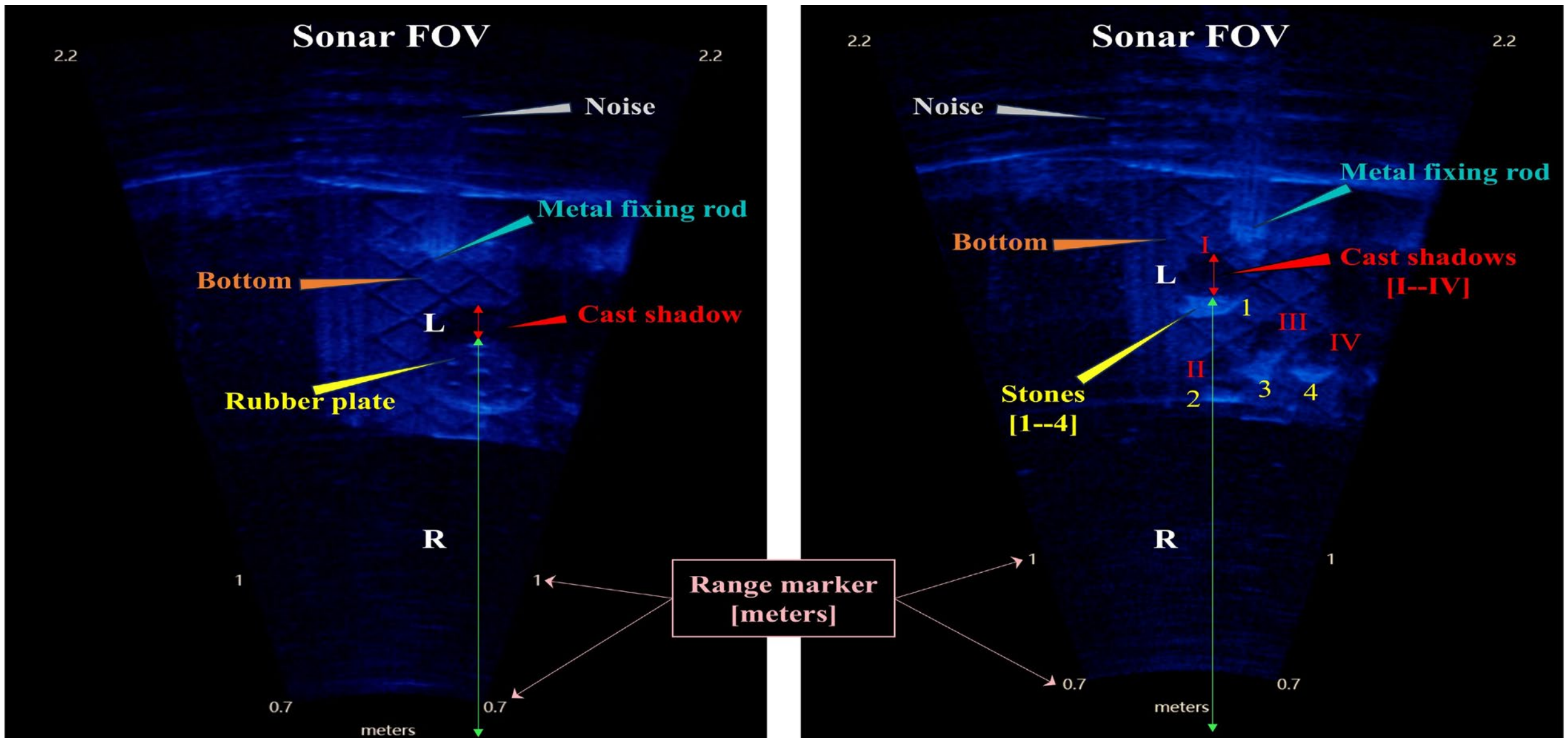

Figure 14. Analysis of visual clues in acoustic camera images.

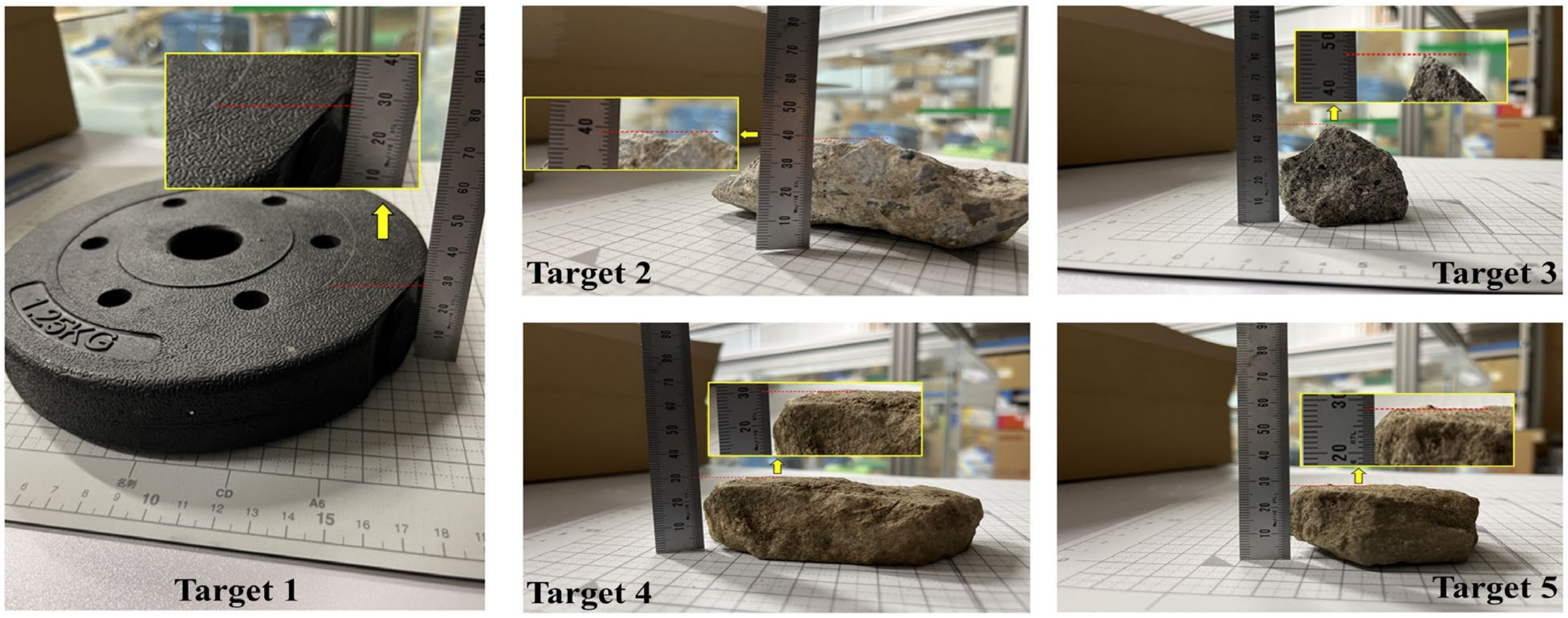

Figure 15. The actual height measurement (ground truth) for each target in the experiment.

In this study, acoustic camera posture adjustments involve two main aspects: firstly, generating altitude variation via simple ascent or descent; and secondly, adjusting the pitch angle of the acoustic camera to optimize the distribution of cast shadows. In actual seafloor surveys, these adjustments in sonar posture are easily achievable. Ascent and descent of the sensor can be finished via the thrusters of the marine robot, while pitch angle adjustments can be achieved with the aid of the rotating device Rotator [4]. For broader investigations, marine robots can be equipped with sonar to scan the seabed environment at various altitudes, as illustrated in Fig. 1. Then, by employing image-matching techniques, image pairs of seabed targets captured at varying altitudes can be obtained. This method will enhance the overall survey efficiency.

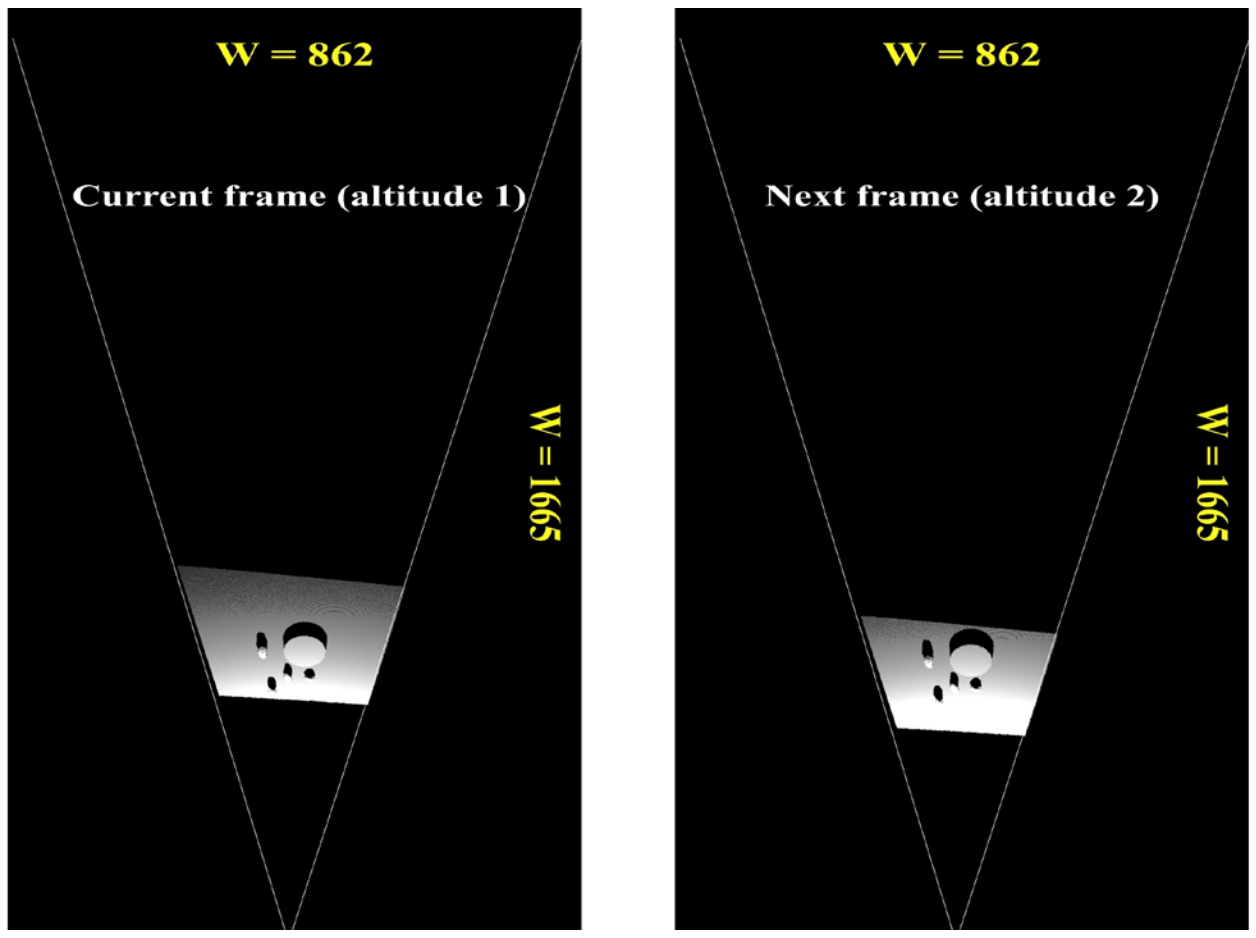

Figure 16. Target acoustic images generated by the sonar simulator.

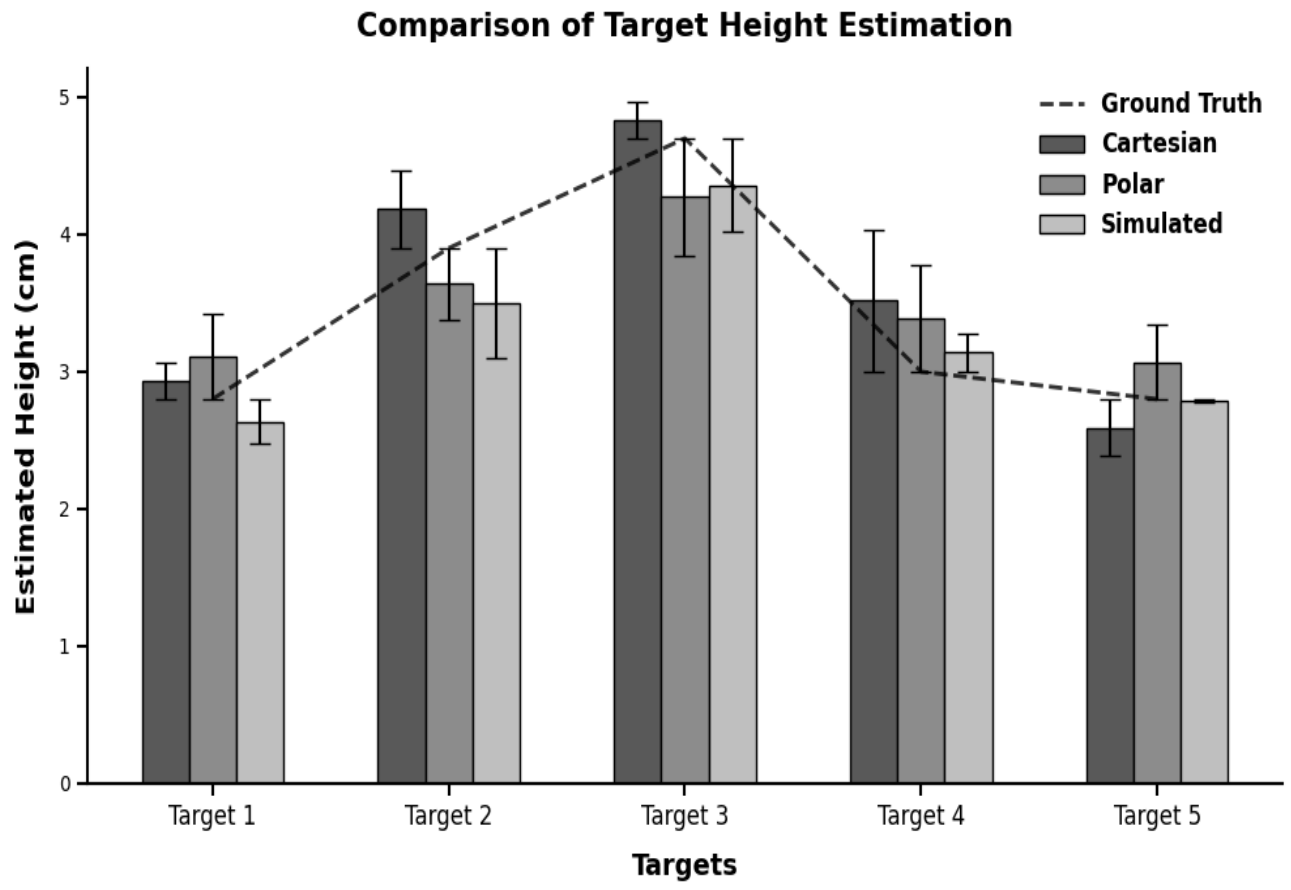

Figure 17. Comparison of target height estimation results.

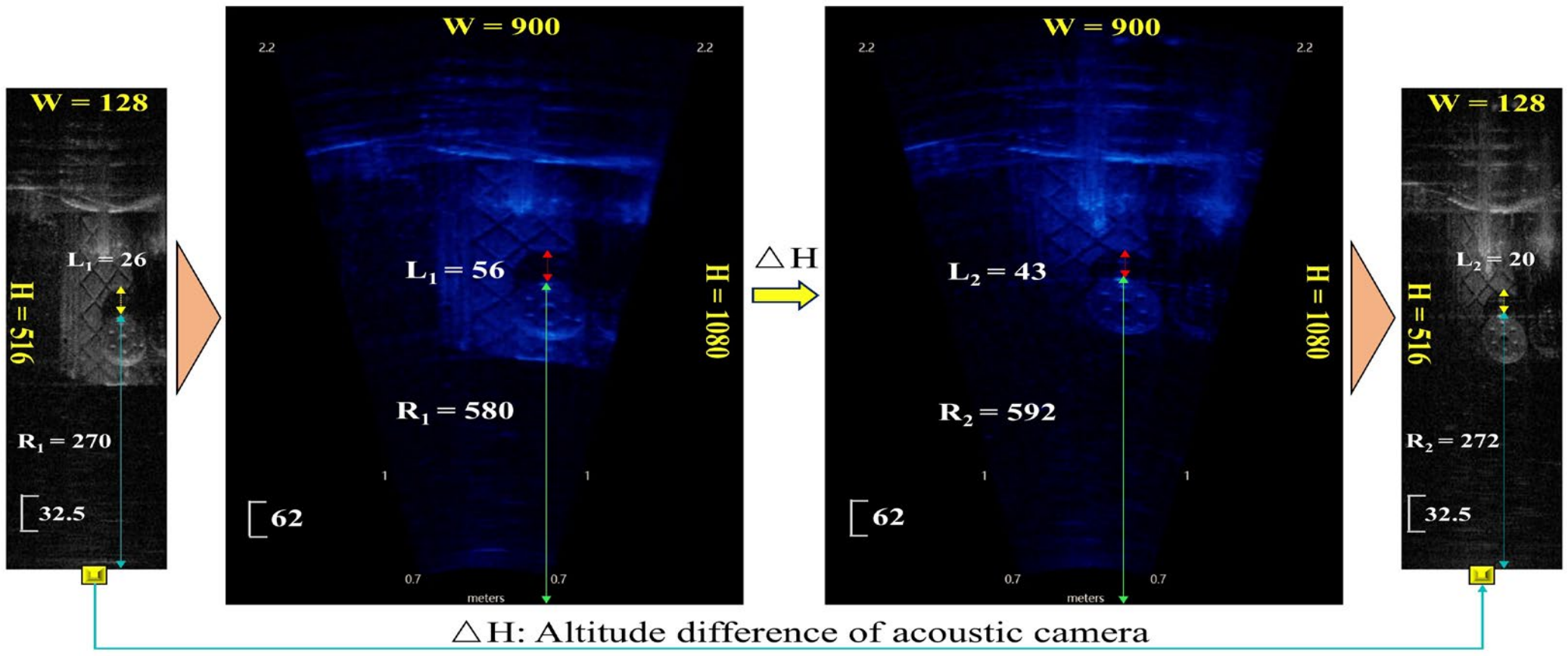

Figure 18. Visualization of target height estimation process.

## V. Results and Evaluation

In this section, we utilize the acoustic camera images collected during the experiments and the recorded sensor posture data to estimate the height of targets. Specifically, we select sonar images captured when the altitude of the acoustic camera changes by 0.1 m for testing. The process of target height estimation is visualized in Fig. 18, and statistical estimation results are presented in Fig. 17 and Table III, where n* represents the rectangular sonar image expressed in polar coordinates, n-S denotes the simulated sonar image, and Est. refers to the estimated target height.

It can be observed that under the current range resolution, the accuracy of target height estimation is high. However, it is important to note that when performing height estimation in actual fan-shaped images, the estimation error tends to be larger because the bottom of the fan-shaped image does not correspond to the sonar zero-range baseline. Therefore, the conversion process of raw data to fan-shaped sonar images can be further optimized to improve the estimation accuracy.

In the acoustic camera images from the water tank, where noise and artifacts complicate accurate shadow boundary extraction, we performed target height estimation in both Cartesian and polar coordinate systems to compare errors, while in the simulated images, with high-quality cast shadow distribution, we directly estimated target height in the Cartesian system and recorded the error.

The survey approach proposed in this work breaks through the traditional seabed mosaicking paradigm, which typically requires scanning at a fixed altitude and constant pitch angle. Our proposal finally achieves both the generation of seabed mosaicking and the acquisition of target height data with minimal increase in operational cost. It enhances the richness of survey data and enables rapid seabed 3D perception.

TABLE III. Target Height Estimation Results

| Target | R1 [pixel] | L1 [pixel] | R2 [pixel] | L2 [pixel] | GT [cm] | Est. [cm] | Error [cm] |
|---|---|---|---|---|---|---|---|
| 1 | 580 | 56 | 592 | 43 | | 2.932 | -0.132 |
| 1* | 270 | 26 | 272 | 20 | 2.8 | 3.110 | -0.310 |
| 1-S | 518 | 28 | 535 | 24 | | 2.637 | +0.163 |
| 2 | 585 | 61 | 587 | 49 | | 4.185 | -0.285 |
| 2* | 291 | 29 | 294 | 23 | 3.9 | 3.639 | +0.261 |
| 2-S | 493 | 32 | 514 | 28 | | 3.500 | +0.400 |
| 3 | 462 | 50 | 475 | 42 | | 4.832 | -0.132 |
| 3* | 219 | 25 | 222 | 20 | 4.7 | 4.274 | +0.426 |
| 3-S | 417 | 26 | 440 | 24 | | 4.358 | +0.342 |
| 4 | 513 | 35 | 525 | 30 | | 3.517 | -0.517 |
| 4* | 248 | 22 | 256 | 18 | 3.0 | 3.390 | -0.390 |
| 4-S | 444 | 26 | 466 | 23 | | 3.141 | +0.141 |
| 5 | 495 | 35 | 504 | 28 | | 2.593 | +0.207 |
| 5* | 242 | 23 | 248 | 18 | 2.8 | 3.071 | -0.271 |
| 5-S | 436 | 20 | 457 | 18 | | 2.786 | +0.014 |

## VI. Discussion

This study simplifies the issue of target height estimation in acoustic camera images into the analysis of the physical phenomenon of target cast shadow and proposes a target height calculation model based on sensor posture adjustment. The proposed approach has been validated through practical experiments and simulation tests, demonstrating its feasibility. However, this approach is still in the early stages of research and has some limitations that need to be addressed.

*1) Seafloor plane simplification:* In this work, we assume a flat seafloor, which aims to verify the approach accuracy. When there are obvious undulations on the seabed surface, the sound wave propagation path will deviate. Therefore, the influence of diffuse reflection of sound waves needs to be fully considered [17]. Multi-path reflection will make the cast shadow blurred and irregular [12], and the introduced noise interference will reduce the accuracy of shadow visual clues extraction, further affecting the accuracy of target height estimation. In underwater survey scenarios requiring high precision, additional calibration steps may be necessary to correct the errors introduced by the flat surface assumption.

*2) Cast shadow segmentation:* In practical seabed surveys, accurately extracting the cast shadow areas of objects from normal reverberation background energy is a challenging image processing step. In this study, we applied a classical adaptive threshold segmentation method. However, this method performs poorly in complex seabed environments, as background noise and reflection artifacts interfere with shadow segmentation, leading to inaccurate results. Currently, we are developing a novel learning-based target cast shadow segmentation model. By training on a dedicated acoustic image dataset, the model could learn more complex feature patterns, enabling it to identify the background and cast shadow, thereby improving the shadow extraction accuracy.

*3) Sensor posture disturbance:* The current height estimation models have not accounted for the impact of sensor posture disturbances. However, in actual seabed surveys, especially in dynamic environments, sensor stability may be affected by external interference, leading to changes in its posture, which directly impact the height estimation. Next, we will conduct water tank experiments and simulator experiments to evaluate the impact of external interference.

## VII. Conclusion

In this study, we propose a new method for estimating the height of seabed targets to compensate for 2D seabed mosaicking. This method infers the target height by analyzing the cast shadow differences caused by the change in the acoustic camera posture. This design pattern does not require additional sensor data and can be implemented using only a single acoustic camera, without the requirement for complex modeling of elevation loss. Dominant seabed mosaicking methods are usually based on a fixed sonar posture, while the method proposed in this work could obtain both 2D seabed panoramas and target 3D height data with almost no increase in operating costs, thereby enriching the survey information.

Future research will involve integrating acoustic cameras into marine robots and validating the proposed method in more complex seafloor environments. Furthermore, this method can be embedded into the visual processing framework of marine robots to further enhance the efficiency of seabed perception.

## Acknowledgment

The authors would like to acknowledge PENTA-OCEAN CONSTRUCTION CO., LTD for their support in providing the acoustic camera ARIS EXPLORER 3000 for this study.